# Lightweight Change Detection in Heterogeneous Remote Sensing Images with Online All-Integer Pruning Training

Chengyang Zhang, Weiming Li, Gang Li, *Senior Member*, *IEEE*, Huina Song, Zhaohui Song, Xueqian Wang, *Member*, *IEEE*, and Antonio Plaza, *Fellow*, *IEEE*

*Abstract*—Detection of changes in heterogeneous remote sensing images is vital, especially in response to emergencies like earthquakes and floods. Current homogenous transformation-based change detection (CD) methods often suffer from high computation and memory costs, which are not friendly to edge-computation devices like onboard CD devices at satellites. To address this issue, this paper proposes a new lightweight CD method for heterogeneous remote sensing images that employs the online all-integer pruning (OAIP) training strategy to efficiently fine-tune the CD network using the current test data. The proposed CD network consists of two visual geometry group (VGG) subnetworks as the backbone architecture. In the OAIP-based training process, all the weights, gradients, and intermediate data are quantized to integers to speed up training and reduce memory usage, where the per-layer block exponentiation scaling scheme is utilized to reduce the computation errors of network parameters caused by quantization. Second, an adaptive filter-level pruning method based on the L1-norm criterion is employed to further lighten the fine-tune process of the CD network. Experimental results show that the proposed OAIP-based method attains similar detection performance (but with significantly reduced computation complexity and memory usage) in comparison with state-of-the-art CD methods.

This work was supported by National Key R&D Program of China under Grant 2021YFA0715201, in part by National Natural Science Foundation of China under Grants 62101303, 62341130 and 62101167, Junta de Extremadura under Grant GR18060, Spanish Ministerio de Ciencia e Innovacion through project PID2019-110315RB-I00 (APRISA), and the European Union's Horizon 2020 research and innovation program under grant agreement No. 734541 (EOXPOSURE) Junta de Extremadura under Grant GR18060, Spanish Ministerio de Ciencia e Innovacion through project PID2019-110315RB-I00 (APRISA), and the European Union's Horizon 2020 research and innovation program under grant agreement No. 734541 (EOXPOSURE). This work has been accepted in part at the IEEE International Geoscience and Remote Sensing Symposium, 2024 [50].

C. Zhang, W. Li, X. Wang, and G. Li are with the Department of Electronic Engineering, Tsinghua University, Beijing 100084, China.

A. Plaza is with the Hyperspectral Computing Laboratory, Department of Technology of Computers and Communications, Escuela Politécnica, University of Extremadura, 10003 Cáceres, Spain.

H. Song and Z. Song are with the Zhejiang Key Laboratory of Space Information Sensing and Transmission, Hangzhou Dianzi University, Hangzhou, 310018, China.

Corresponding author: Xueqian Wang (Email: wangxueqian@mail.tsinghua.edu.cn)

## I. INTRODUCTION

CHANGE detection (CD) aims to detect changed regions on the earth using bi-temporal remote sensing images. Efficient CD is essential for disaster rescue tasks (e.g., earthquakes and floods), as it can timely provide information about the disaster areas. CD using homogenous remote sensing images has been widely investigated for many years [1]-[3], [48], [49]. This includes, for example, bi-temporal images that are both collected by optical satellites. Homogeneous bi-temporal images have similar feature space, which makes them convenient to design CD algorithms. In practice, because of the difficulty of collecting homogeneous remote sensing images on time for urgent change events [12], and due to the robustness derived by the complementary use of image data from different imaging modalities, CD based on heterogeneous remote sensing images is a practical alternative in emergency cases. However, the significantly different features of heterogeneous images [such as optical and synthetic aperture radar (SAR) images] make CD methods designed for homogeneous images [1]-[3], [48], [49] unfeasible for heterogeneous image cases.

Recently, homogeneous transformation has been a popular strategy for CD using heterogeneous remote sensing images. The homogeneous transformation aims to transform heterogeneous images into a homogeneous feature space to compare them directly and perform CD in that space. The superiority of homogeneous transformation lies in its intuitiveness and universal applicability [12]. Homogeneous transformation-based CD methods can be roughly classified into two categories, i.e., hand-crafted feature space-based methods and neural network-based methods.

The aim of hand-crafted feature space-based homogeneous transformation methods is to learn specific feature parameters and then transform the heterogeneous remote sensing images to the hand-crafted feature space, where heterogeneous images are comparable and changed/unchanged areas can be distinguished. The work in [4] proposes a coupled dictionary learning method to jointly learn sparse representation space of unchanged samples. In this way, change pixels of heterogeneous images exhibit larger reconstruction errors than unchanged ones in the sparse representation space, and the former ones are effectively highlighted. The homogeneous pixel transformation (HPT) [5] method utilizes the position and similarity information of *K*-nearest neighboring pixels of the current pixel in one heterogeneous image to reconstruct the corresponding pixel in the other image. The larger the difference between the reconstructed and original pixels means the higher the probability of change. The work in [6] employs the affinity matrix comparison approach to online select unchanged pixels and uses regression methods to achieve the homogeneous transformation of heterogeneous images. In [7], a patch-level *K*-nearest neighbor algorithm-based CD method is proposed to measure the structure consistency of heterogeneous images and low consistency regions therein tend to be identified as changed areas. Note that patch-level feature homogeneous transformation space [7] is more robust than pixel-level ones [5] when heterogeneous images are contaminated by noisy pixels.

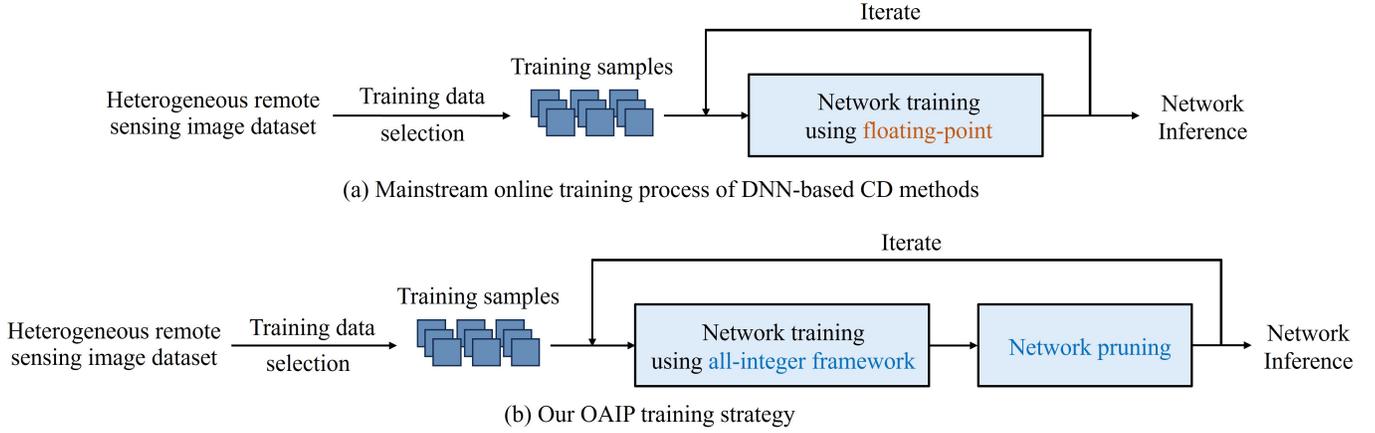

Fig. 1. Comparison between (a) the mainstream online training process of DNN-based CD methods and (b) our proposed OAIP-based CD method.

Deep neural network (DNN)-based CD methods are gaining popularity for heterogeneous remote sensing images with the development of deep learning. DNN-based homogeneous transformation CD approaches use neural networks to transform heterogeneous images into a homogeneous feature space while preserving their semantic information. Due to the lack of massive training datasets for offline training, existing training strategies for DNN-based CD methods in heterogeneous remote sensing images are often based on online training [8]-[11]. During online training, selected samples from the current test image data are utilized to fine-tune pre-trained DNN models, and the fine-tuned models are then used to infer the entire test dataset. In [8], a symmetric structured Siamese network is designed for the homogeneous transformation of heterogeneous images, and a pre-training strategy is developed to reduce the influence of image noises. The work in [11] uses two visual geometry group (VGG) subnetworks as style extractors to transform the style of the SAR images to that of the optical images while maintaining the semantic content to more easily perform CD. Works [9] and [10] propose generative adversarial network-based CD methods, where neural networks are employed as generators for homogeneous transformation and image patches labeled by the affinity matrix comparison strategy are utilized as online training samples.

Although heterogeneous CD approaches [8]-[11] have achieved remarkable CD performance, they often suffer from substantial computational complexity and large memory usage. The main reason is that, although online training using samples from the current test dataset is fine-tuned instead of training from scratch [12], the online fine-tuning process involves learning and updating lots of parameters for many iterations, which degrades the efficiency of the heterogeneous CD process. Recently, several research efforts have been devoted to investigating onboard data processing for remote sensing applications due to its low information latency [33], [34]. However, onboard data processing is based on edge-computation devices that often do not have enough resources [35] to support the heterogeneous CD with high computation and memory costs.

Several existing methods have been developed to reduce the iterations and online training time of CD networks using heterogeneous remote sensing images. The work in [30] introduces a loss function with focal and dice loss terms to address the impact of imbalanced positive and negative samples during online training, resulting in fast convergence of online training and reduced computational complexity. In [36], the parameters in the heterogeneous CD network are updated with an adaptive learning rate to achieve rapid convergence. The hierarchical attention feature fusion method developed by [37] enhances the efficiency of CD networks by concentrating on high-quality semantic content during online training. Note that the aforementioned methods [30], [36], [37] still face two issues in terms of efficient CD in heterogeneous images. 1) Existing methods are often predominantly based on floating-point computations, which are slower and not hardware-friendly when compared with integer-based calculations. 2) In the online training process, large amounts of neural network parameters need to be trained, leading to significant redundancy in calculations. These two drawbacks still limit the deployment of existing methods on edge computation platforms with limited computation and memory resources.

In this paper, we propose a new lightweight CD method for heterogeneous remote sensing images (see Fig. 1), where a novel online all-integer pruning (OAIP) training strategy is developed to quantize the network parameters to integers and prune redundant parameters during online training. Inspired by works [11] and [12], two VGG subnetworks are used as our foundational CD network architecture, and a patch-based affinity matrix comparison method is employed to obtain the training samples for the online training process. During the OAIP training process of the CD network, we first convert all the floating-point operations into all-integer operations, including convolution, pooling, and normalization operations to reduce the computation and memory costs of online training process of the CD network. Note that training the CD network using all-integer arithmetic is challenging as integers lack dynamic range and precision, leading to large computation errors when computing network parameters. These errors accumulate during online training, resulting in slower model convergence and degraded CD performance. To address this issue, we utilize the per-layer block exponentiation scaling scheme (PLBESS) to reduce the errors caused by integer calculations and improve the online training performance. Second, we develop an adaptive filter-level pruning strategy based on the L1-norm criterion to gradually trim the backbone network during the online training process. Experimental results with heterogeneous remote sensing images show that our newly proposed method provides better CD efficiency while maintaining CD performance in comparison with existing state-of-the-art heterogeneous CD methods.

The rest of this paper is organized as follows. The existing lightweight heterogeneous CD methods, as well as related works concerning integer training and network pruning are described in Section II. The heterogeneous CD problem is formulated in Section III. Section IV elaborates the proposed OAIP-based method for CD in heterogeneous remote sensing images. In Section V, a series of in-depth experiments on three datasets are discussed to demonstrate the advantages of the proposed method. Section VI concludes this paper with some remarks and hints at plausible future research.

## II. RELATED WORK

This section starts with a description of existing lightweight DNN-based CD methods. Subsequently, we introduce prevalent integer training and pruning methods for DNNs, which are commonly used to lighten the training process. Finally, we clarify the difference between our proposed OAIP-based method and other existing methods.

### A. DNN-Based Lightweight CD Methods

Existing DNN-based lightweight CD methods primarily involve manually designing concise networks or lightweight online training strategies to reduce the number of parameters that need updating/fine-tuning. A commonly used lightweight approach is to employ the efficient U-shaped network (UNet) as feature extractors [38] or homogeneous transformation functions [39], [40] to reduce the computation burden of the online training process of CD networks. The work in [41] first utilizes an efficient autoencoder network to generate pseudo-labels of the partial test data, which are then used to train a fully connected network to achieve lightweight CD. The work in [42] achieves feature extraction and homogeneous feature space mapping using an efficient multi-layer forward encoding network to jointly learn semantic features during online training. The CD method in [12] only learns the weights of deep convolutional layers in VGG networks, while low-level layer parameters of VGG are directly transferred from VGG trained by ImageNet [44] and frozen during the online training process. This reduces the network parameters to be online updated. A common characteristic of the methods in [12] and [38]-[42] is that they reduce the number of training parameters compared with baseline methods, leading to a lightweight CD process. However, these lightweight methods in [12], [38]-[42] do not consider the redundancy of network parameters to be updated during the online training process. Additionally, they overlook the limited computational and memory efficiency of floating-point calculations.

### B. Integer Training and Pruning Methods

Integer training in neural networks refers to representing the data (partially or fully) using integers instead of floating-point numbers during training. This aims to reduce model memory usage and accelerate the training process. The research on integer training can be traced back to the inference using low-precision neural networks [13], [14]. Subsequently, researchers attempted to use integer precision during backward propagation for gradient computation to accelerate the training process. The work in [15] quantizes the weight gradients to 8-bit integers and proposes a gradient clipping method to reduce quantization errors during the network training. In [16], the forward propagation, backward propagation, and weight updating are operated using 16-bit integer precision and multiple rounding methods are utilized to accelerate the training process of the network and maintain detection performance. In [17], during the network training process, weights, activations, and gradients are all quantized to integers and both convolution and normalization operations are formulated into integer-based operations to achieve lightweight training performance. Even though the aforementioned integer training methods primarily compute gradients of parameters using integers, a number of floating-point calculations are still employed throughout the entire training process. For instance, in [16] and [17], the floating-point precision format is used to store the accumulated results of integer-based convolution to avoid data overflow, while in [15] the floating-point precision format is still employed to calculate the learning rate to obtain accurate detection results. To avoid the floating-point calculations, the work in [23] proposes an all-integer network training framework, which only uses integers and integer-based calculations during online training, achieving a more lightweight training process.

Neural network pruning refers to the process of removing unimportant weights during training, either one shot or multiple shots, aiming to reduce both the inference and training computation complexity. To facilitate the deployment of DNNs on hardware platforms and achieve acceleration, researchers have established structured pruning methods [18]-[20] that regularly prune groups of weights (in the filter and channel dimension) to achieve lightweight performance. The work in [18] introduces trainable scaling factors to represent the significance of filters during training and prunes the filters with small significance factors. In [19], scaling factors of the batch normalization (BN) layer are used to reflect channel importance allowing for the pruning channels to accelerate training. The work in [20] prunes filters in the network with small L1-norm values to make the network more efficient. Although the structured pruning methods (based on the simple L-norm criterion or trainable factors) are easy to be deployed on hardware, they are prone to causing performance loss [32]. To alleviate this issue, adaptive pruning methods have been proposed. In [21], filters are classified into unimportant and important ones, and the former are trained with the L1 regularization term and pruned by an adaptive threshold in the training process. A trainable pruning threshold is adjusted during training to adaptively achieve the optimal pruning performance in [22]. In the eager pruning strategy [24], the network weights are pruned greedily and an adaptive rolling back operation is performed when the important weights are pruned to maintain the detection performance. The adaptability of pruning in [21]-[22], [24] enables them to reduce performance degradation with fewer network parameters.

### C. Differences between Our OAIP and Existing Works

To the best of our knowledge, current CD methods in heterogeneous remote sensing images have yet to leverage pruning or integer training techniques for accelerating online training and enhancing CD efficiency. In comparison with existing heterogeneous CD methods [38]-[42], we substitute all-floating-point calculations with all-integer computations and employ the adaptive pruning method, effectively lightening the entire training process. In comparison with existing lightweight DNN-based methods with integer training and pruning [20], [23], our newly proposed method further combines all-integer training with adaptive pruning, resulting in a more lightweight training process without noticeable performance loss. As opposed to the fine-grained pruning method in [24], our approach employs filter-level pruning for smooth acceleration and hardware deployment.

### III. PROBLEM FORMULATION

We define $I^{pre} \in \mathbb{R}^{H \times W \times C_1}$ and $I^{post} \in \mathbb{R}^{H \times W \times C_2}$ as two heterogeneous remote sensing images before and after the change event, where $H$ and $W$ denote the image height and width, respectively, and $C_1$ and $C_2$ represent the numbers of channels of $I^{pre}$ and $I^{post}$, respectively. Those two images are co-registered and rescaled with each other before the CD processing. Note that image registration is not the focus of our paper and some advanced image registration methods for heterogeneous remote sensing images can be found in [46], [47]. The objective of CD is to identify the change areas between $I^{pre}$ and $I^{post}$ and generate a binary change map $I_b$, where "1" and "0" represent changed and unchanged areas, respectively.

Based on the widely adopted homogeneous transformation strategy [8], [12], the model of change detection in heterogeneous remote sensing images can be formulated as

$$I_b = f_b(I_d) \,, \tag{1}$$

$$I_d = f_d(t_1(I^{pre}), t_2(I^{post})) \,, \tag{2}$$

where $t_1$ and $t_2$ represent two homogeneous transformation functions that map the two images into the same feature space, and $f_d$ is a difference operation that compares the features extracted by $t_1$ and $t_2$ to generate a difference map $I_d$. $f_b$ in Eq. (1) is a threshold segmentation function that segments $I_d$ into the final binary change detection map $I_b$.

### IV. PROPOSED METHOD

In this section, we first introduce a patch-level affinity matrix comparison-based training sample selection method. Then, the Siamese VGG structured CD network and the proposed efficient OAIP-based training strategy are illustrated.

### A. Selection of Training Samples

The online training process requires the training samples selected from the current bitemporal images to be detected. In this subsection, we develop a patch-level affinity matrix comparison method inspired by [9] to select training samples for OAIP. First, the heterogeneous image pair { $I^{pre}$, $I^{post}$ } is uniformly co-divided into patches with $s \times s$ size, i.e., { $I_i^{pre}$, $I_i^{post}$ }, $i = 1, 2, \ldots, Q$, where $Q$ denotes the quantity of patches in each heterogeneous image. Let $A_{i,j}^{pre}$ and $A_{i,j}^{post}$ denote the value in the $i$th row and $j$th column of affinity matrices of $I^{pre}$ and $I^{post}$, respectively, where

$$A_{i,j}^{\tau} = \exp\left\{-\frac{(d_{i,j}^{\tau})^2}{h_{\tau}^2}\right\}, \tag{3}$$

$$d_{i,j}^{\tau} = \left| M_i^{\tau} - M_j^{\tau} \right|, \tag{4}$$

$\tau \in \{pre, post\}$, $M_i^{pre}$ and $M_i^{post}$ represent the mean value of $I_i^{pre}$ and $I_i^{post}$, respectively. The kernel width $h_{\tau}$ in Eq. (3) is adaptively determined by

$$h_{\tau} = \max\left\{d_{i,j}^{\tau}\right\}, \quad i, j \in \{1, \ldots, Q\}. \tag{5}$$

A larger value of $A_{i,j}^{\tau}$ means that the $i$th and $j$th patches in an image are more similar. Based on $A_{i,j}^{pre}$ and $A_{i,j}^{post}$, the change score of the $i$th patch (pair) in the bitemporal images is calculated by

$$\alpha_i = \sum_{j=1}^{Q} \left| A_{i,j}^{pre} - A_{i,j}^{post} \right|, \quad i \in \{1, \ldots, Q\}. \tag{6}$$

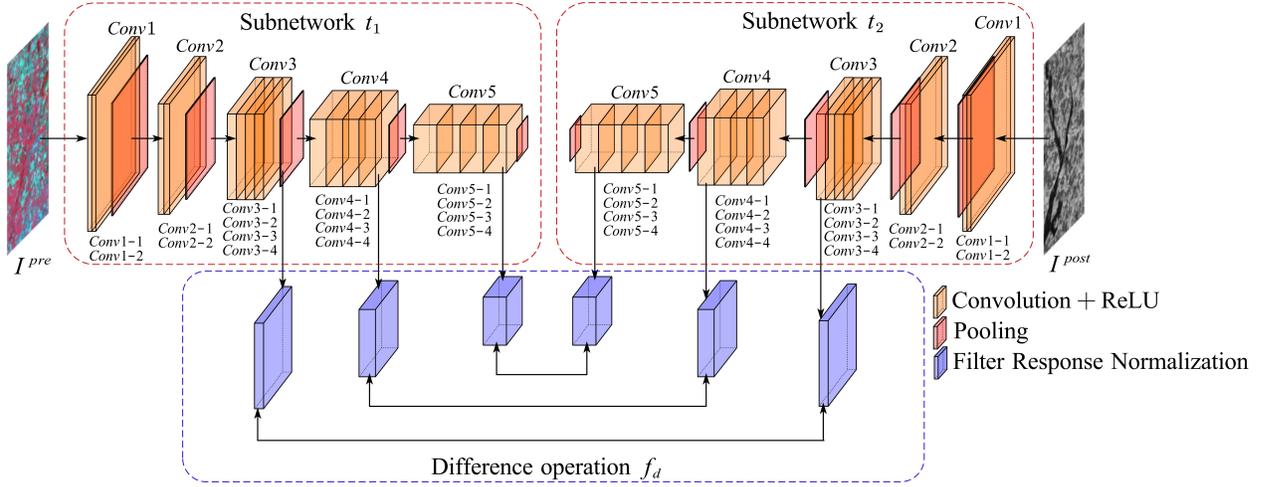

Fig. 2. Overall structure of our CD network.

Then, we select $n_h$ patch pairs with the highest change scores (positive samples) and $n_l$ patch pairs with the lowest change scores (negative samples) to fine-tune the CD network. Section V.C will show the effect of $n_h$ and $n_l$ on the CD performance.

It is worth mentioning that our training sample selection method is a more efficient version of that in [9], which computes pixel-level affinity matrices with the computational complexity $HW / s^2$ times greater than that of our patch-level affinity matrices. In addition, patch-level sample selection demonstrates enhanced robustness to image noise compared with pixel-level method in [9].

### B. Network Architecture and Loss Function

Similar to most CD network structures [9]-[12], our CD network utilizes the Siamese VGG structure network that transforms heterogeneous images into the same homogenous feature space while maintaining semantic information. The structure of our CD network is shown in Fig. 2.

Our CD network has three parts, i.e., two Siamese VGG subnetworks standing for homogenous transformation function $\{t_1, t_2\}$ in Eq. (2), respectively, and a difference operation $f_d$ for CD result generation. In Fig. 2, *Conv* represents the convolutional layer. Each subnetwork comprises 16 convolutional layers, denoted as *Conv* 1-1 to *Conv* 5-4. The weights of *Conv* in our CD network are pre-trained using the Cifar-10 dataset [29]. Following [12], only the weights of $\Theta = \{Conv\ m\text{-}4, m=3, 4, 5\}$, are fine-tuned in the online training process to reduce the computation complexity. The outputs of $\Theta$ are used as the input of difference operation $f_d$ in Eq. (2) to compute the difference map $I_d$ because they represent deep-level semantic content features and are beneficial to highlight change areas [12].

Due to the variations of outputs of the $\Theta$, it is necessary to normalize the semantic content features produced by these layers. Instead of the BN layers to normalize the outputs [12], our CD network applies the L1-norm filter response normalization (FRN) layer [26] to normalize the outputs of $\Theta$ because of its efficiency and robustness for small-batch training [25].

Based on our VGG network architecture and training sample selection strategy, we design the loss function $\mathcal{L}$ to train our CD network:

$$\mathcal{L} = \mathcal{L}_- + \mathcal{L}_+ , \qquad (7)$$

$$\mathcal{L}_- = \sum_{m=3}^{5} \left[ \frac{\alpha_m}{W_m H_m C_m} \sum_{x=1}^{W_m} \sum_{y=1}^{H_m} \sum_{z=1}^{C_m} (\text{FRN}_{x,y,z}^{t_1, m, -} - \text{FRN}_{x,y,z}^{t_2, m, -})^2 \right], \quad (8)$$

$$\mathcal{L}_+ = -\sum_{m=3}^{5} \left[ \frac{\alpha_m}{W_m H_m C_m} \sum_{x=1}^{W_m} \sum_{y=1}^{H_m} \sum_{z=1}^{C_m} (\text{FRN}_{x,y,z}^{t_1, m, +} - \text{FRN}_{x,y,z}^{t_2, m, +})^2 \right], (9)$$

where $\mathcal{L}_-$ and $\mathcal{L}_+$ are sub-loss functions using negative samples and positive samples, respectively. $\text{FRN}_{x,y,z}^{t_p, m, \upsilon}$ is the output of the FRN that is applied after $\Theta$ of subnetwork $t_p$ ($p = 1, 2$). $\upsilon = +, -$ means positive and negative samples, respectively. The subscript $(x, y, z)$ in Eqs. (8) and (9) denote the index of elements of $\text{FRN}^{t_p, m, \upsilon}$. $H_m$, $W_m$, and $C_m$ represent the height, width, and number of channels of the $\text{FRN}^{t_p, m, \upsilon}$, respectively. $\{\alpha_m, m=3, 4, 5\}$ are constants that control the influence of the outputs of $\{Conv\ m\text{-}4, m=3, 4, 5\}$.

### C. Network Training with Online All-Integer-Pruning

Before the CD inference stage, we online train (fine-tune) the Siamese VGG network using pre-selected samples in Section IV.A. Prior to the online training, all the floating-point weights in two pre-trained VGG sub-networks, as well as the input data, are uniformly quantized to 8-bit integers, which is a widely used precision in network quantization.

Algorithm 1 shows the whole training process of the proposed method. During training, each iteration can be divided into two parts [see Fig. 1(b)], i.e., the all-integer weight learning part and the pruning part. The all-integer

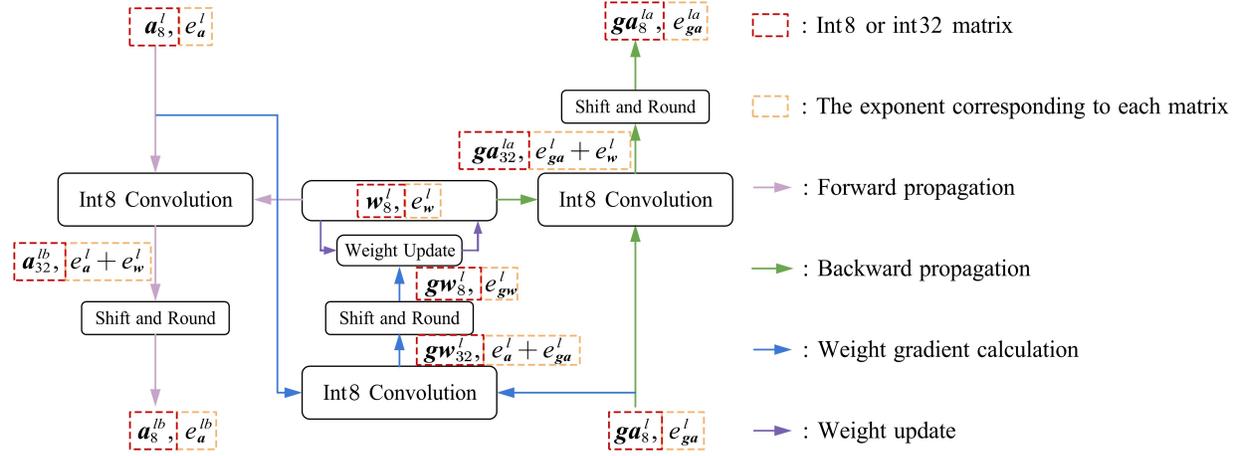

Fig. 3. The forward propagation, backward propagation, weight gradient calculation, and weight update processes.

---

**Algorithm 1** OAIP Training Process

**Input:** two quantized subnetworks, training samples, iteration interval of pruning $P$ (even number), the maximum pruning rate $r^{\max}$ in each pruning operation, total training iterations $T$, and tolerated percentage threshold $\theta$.

**Output:** the trained CD network with int8 weights and fewer filters.

1   $N^l = r^{\max} \times N_e^l$, $\forall l$, $l \in \Theta_p$.
2   **for** $k = 1, 2, ..., T$ **do**
        // All-integer weight learning part
3       Forward propagation based on PLBESS using training samples.
4       Calculate loss function value $\mathcal{L}_k$.
5       Backward propagation based on PLBESS.
6       Weight gradient calculation based on PLBESS.
7       Update the int8 weights of $\Theta$.
        // Pruning part
8       **if** $k \bmod P = 0$ **do**   //Pruning operation
9           Prune $N^l$ filter pairs in layer $l$, $\forall l$, $l \in \Theta_p$.
10      **end if**
11      **if** $k \bmod P = P / 2$ **do**   //Checking rolling back.
12          Let $\mathcal{L}^{\max} = \max\{\mathcal{L}_q, q = 1, 2, ..., k\}$ and $\mathcal{L}^{\text{last}} = \mathcal{L}_{k-P/2-1}$ be the loss value just before the last pruning.
13          **if** $\mathcal{L}^{\max} - \mathcal{L}_k < \theta(\mathcal{L}^{\max} - \mathcal{L}^{\text{last}})$ **do**
14              $N^l = N^l / 2$, $\forall l$, $l \in \Theta_p$.
15              $k = k - P / 2$ and roll back the structure and parameters of network just before the last pruning.
16          **else do**
                //Attempt greedy pruning at the next pruning.
17              $N^l = r^{\max} \times N_e^l$, $\forall l$, $l \in \Theta_p$.
18          **end if**
19      **end if**
20  **end for**

---

weight learning part encompasses forward and backward propagation, loss calculation, weight gradient calculation and weight update. The pruning part includes pruning operation and rolling back operation.

*1) All-Integer Weight Learning Part*

The core of the forward propagation, backward propagation and weight gradient calculation during the training of our network is the convolution operation

$$a^{lb} = w^l * a^l \text{ (forward propagation)}, \quad (10)$$

$$ga^{la} = w^l * ga^l \text{ (backward propagation)}, \quad (11)$$

$$gw^l = a^l * ga^l \text{ (weight gradient calculation)}, \quad (12)$$

where the $a$, $w$, $ga$, and $gw$ mean the activation, weight, gradient of activation, and gradient of weight, respectively. The superscripts $l$, $la$, and $lb$ denote the matrix corresponding to layer $l$, the layer above $l$, and the layer below $l$, respectively. Note that we remove bias to make the convolution more efficient in Eqs. (10)-(12). To achieve lightweight performance, int8 precision matrices $a_8^l$, $w_8^l$, $ga_8^l$ and $gw_8^l$ are utilized to approximately represent the full-precision convolution matrices $a^l$, $w^l$, $ga^l$, and $gw^l$, respectively, where the subscripts "8" mean the int8 precision. However, relying solely on int8 precision for convolution causes significant computation errors due to the limited value range of int8. To address this issue, we exploit PLBESS to reduce the errors caused by int8 quantization [23], which assigns int8 numbers $e_a$, $e_w$, $e_{ga}$, and $e_{gw}$ as exponents to each integer matrix in $\{a_8^l, w_8^l, ga_8^l$ and $gw_8^l\}$. The exponent $e$ represents a shared scaling factor for all integers in a matrix. By using PLBESS, the $a^l$, $w^l$, $ga^l$, and $gw^l$ are approximately represented by

$$a^l \approx a_8^l \odot 2^{e_a^l}, \quad (13)$$

$$w^l \approx w_8^l \odot 2^{e_w^l}, \quad (14)$$

$$ga^l \approx ga_8^l \odot 2^{e_{ga}^l}, \quad (15)$$

$$gw^l \approx gw_8^l \odot 2^{e_{gw}^l}. \quad (16)$$

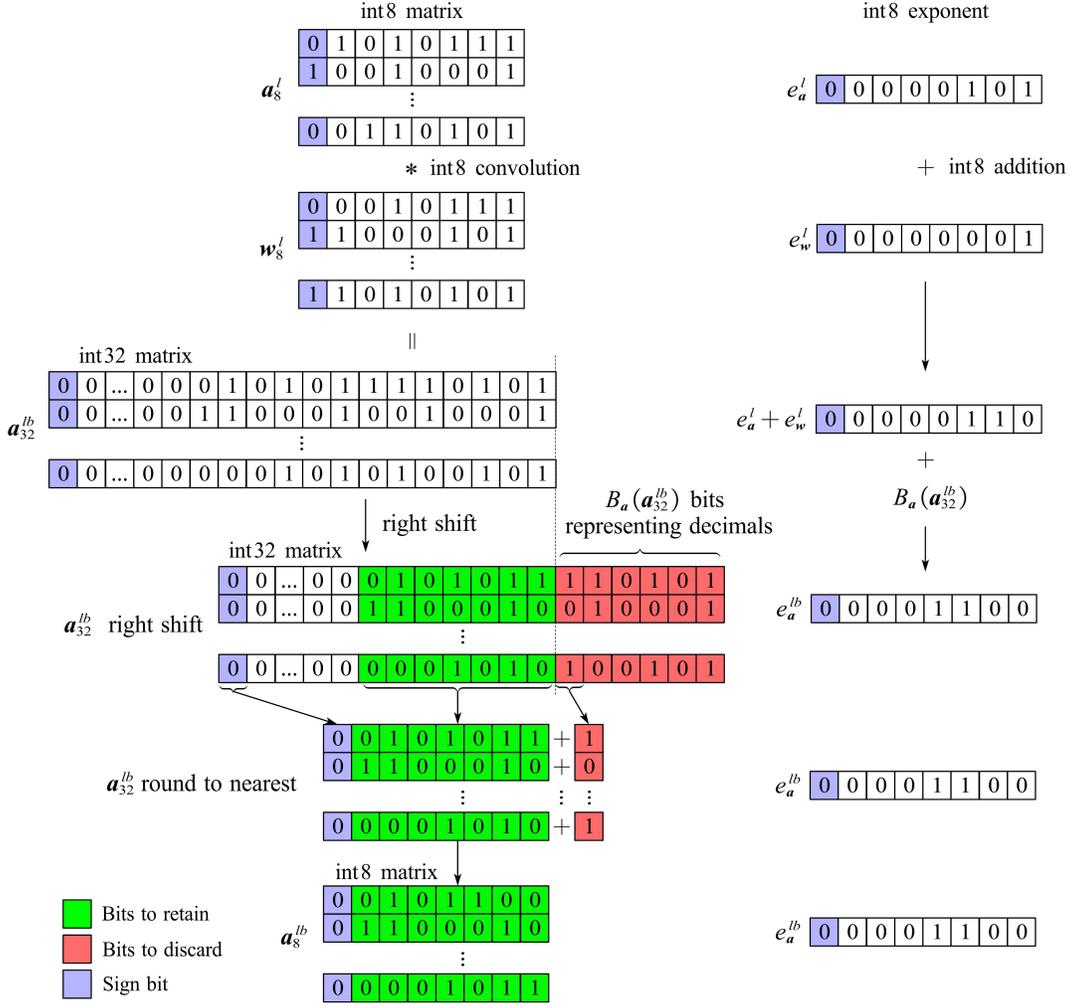

Fig. 4. Int8 convolution and shift-and-round operation flow chart, taking forward propagation as an example.

PLBESS expands the representation range of int8 precision, enabling a more accurate representation of numbers beyond the int8 range, thereby reducing the computational errors caused by int8 precision calculation. Based on Eqs. (13)-(15), the convolution operations in Eqs. (10)-(12) are converted to int8 convolution operations

$$\boldsymbol{a}_{32}^{lb} = \boldsymbol{w}_8^l * \boldsymbol{a}_8^l, \tag{17}$$

$$\boldsymbol{ga}_{32}^{la} = \boldsymbol{w}_8^l * \boldsymbol{ga}_8^l, \tag{18}$$

$$\boldsymbol{gw}_{32}^l = \boldsymbol{a}_8^l * \boldsymbol{ga}_8^l, \tag{19}$$

where the subscript "32" denotes the int32 precision. Here, int32 precision is employed to temporarily store the accumulated results of int8 convolution to avoid data overflow. The exponents of $\boldsymbol{a}_{32}^{lb}$, $\boldsymbol{ga}_{32}^{la}$ and $\boldsymbol{gw}_{32}^l$ are the sum of the exponents of the two input int8 matrices $e_a^l + e_w^l$, $e_{ga}^l + e_w^l$ and $e_{ga}^l + e_a^l$, respectively. Fig. 3 shows the forward propagation, backward propagation, and weight gradient calculation processes employing the int8 convolutions in Eqs. (17)-(19). During the forward propagation, Eq. (17) is performed to obtain the accumulated result $\boldsymbol{a}_{32}^{lb}$. As int8 convolution takes two int8-precision matrices as input, we need to shift-and-round int32 data to int8 before proceeding to the next layer. The details of int8 convolution and shift-and-round operation are depicted in Fig. 4, taking forward propagation as an example. During shift-and-round operation, the effective bit-width is first calculated as:

$$E(\boldsymbol{a}_{32}^{lb}) = \left\lceil \log_2(\max_{x,y,z}(\left|\boldsymbol{a}_{32,x,y,z}^{lb}\right|)) \right\rceil, \tag{20}$$

where "$\lceil \ \rceil$" denotes rounding up operation. The effective bit-width $E(\cdot)$ represents the smallest number of bits required to represent the maximum absolute value in a matrix. The number of bits that need to be discarded is computed as

$$B_a(\boldsymbol{a}_{32}^{lb}) = \max(E(\boldsymbol{a}_{32}^{lb}) - 7, 0), \tag{21}$$

where "7" represents the effective bit-width of the shift-and-round result $\boldsymbol{a}_8^{lb}$. As shown in Fig. 4, all the values in $\boldsymbol{a}_{32}^{lb}$ are then right-shifted $B_a(\boldsymbol{a}_{32}^{lb})$ bits to ensure that their absolute values are all smaller than $2^7$, indicating that they can all be fully represented by the int8 precision. The operation of right-shifting $B_a(\cdot)$ bits is approximately equivalent to a division by $2^{B_a(\cdot)}$. Therefore, we compensate for $B_a(\boldsymbol{a}_{32}^{lb})$ in the exponent part of $\boldsymbol{a}_{32}^{lb}$ to counteract the reduction in value caused by right-shifting and discarding $B_a(\boldsymbol{a}_{32}^{lb})$ bits

$$e_a^{lb} = e_a^l + e_w^l + B_a(\boldsymbol{a}_{32}^{lb}) \,. \tag{22}$$

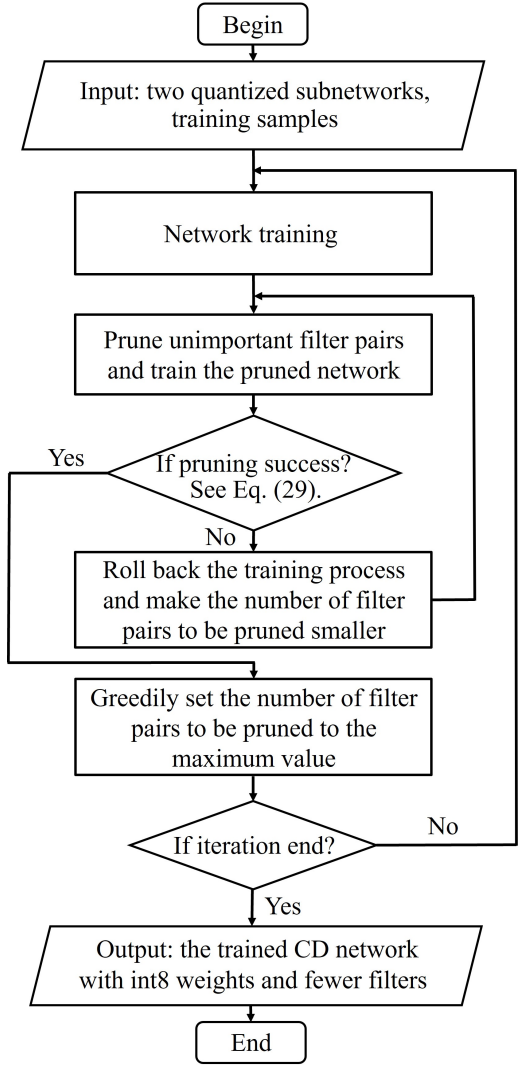

Fig. 5. The flow chart of the adaptive pruning strategy.

See Fig. 4, the $B_a(\boldsymbol{a}_{32}^{lb})$ bits at the end of each value represent a decimal smaller than 1 that needs to be discarded. However, discarding the $B_a(\boldsymbol{a}_{32}^{lb})$ bits at the end directly would introduce significant computational errors. Therefore, for each value in $\boldsymbol{a}_{32}^{lb}$, we employ the round-to-nearest operation to determine whether the decimal represented by $B_a(\boldsymbol{a}_{32}^{lb})$ bits at the end is closer to 0 or 1, and then 0 or 1 is added to the final result for better computation precision. The round-to-nearest operation is a widely used and computationally efficient method for rounding decimals. In Fig. 4, the round-to-nearest operation uses the highest bit in $B_a(\boldsymbol{a}_{32}^{lb})$ bits at the end to determine whether the decimal is closer to 0 or 1. After the shift-and-round operation, we obtain $\boldsymbol{a}_8^{lb}$ and $e_a^{lb}$, which are the inputs of the layer below during the forward propagation. After the forward propagation and loss calculation in Eq. (7), the backward propagation is performed as depicted in Fig. 3. Similar to forward propagation, the convolution in Eq. (18) is conducted during backward propagation to derive accumulated result $\boldsymbol{ga}_{32}^{la}$, and then the right-shifting

$$B_{ga}(\boldsymbol{ga}_{32}^{la}) = \max(E(\boldsymbol{ga}_{32}^{la}) - 7, 0) \tag{23}$$

bits and the round-to-nearest operation are employed to obtain the $\boldsymbol{ga}_8^{la}$. The exponent of $\boldsymbol{ga}_8^{la}$ is

$$e_{ga}^{la} = e_{ga}^l + e_w^l + B_{ga}(\boldsymbol{ga}_{32}^{la}) \,. \tag{24}$$

Subsequently, the convolution of Eq. (19) is carried out to obtain the int32 weight gradient $\boldsymbol{gw}_{32}^l$. The number of bits to right-shift for $\boldsymbol{gw}_{32}^l$ is

$$B_{gw}(\boldsymbol{gw}_{32}^l) = \max(E(\boldsymbol{gw}_{32}^l) - b_{gw}, 0) \,, \tag{25}$$

where $0 < b_{gw} \leq 7$ is a hyperparameter that controls the learning rate of the online training process. Then, the right-shifting operation and round-to-nearest operation are performed to convert $\boldsymbol{gw}_{32}^l$ to $\boldsymbol{gw}_8^l$. The exponent of $\boldsymbol{gw}_8^l$ is

$$e_{gw}^l = e_a^l + e_{ga}^l + B_{gw}(\boldsymbol{gw}_{32}^l) \,. \tag{26}$$

Finally, all the weights are updated and scaled to the int8 precision

$$\boldsymbol{w}_8^{l,k+1} = (\boldsymbol{w}_8^{l,k} - \boldsymbol{gw}_8^{l,k}) \odot \frac{127}{\max\limits_{x,y,z} \left| \boldsymbol{w}_{8,x,y,z}^{l,k} - \boldsymbol{gw}_{8,x,y,z}^{l,k} \right|} \,, \tag{27}$$

where the $k$ means the matrix computed in the $k$th iteration. Based on PLBESS, the errors in the computation of integer training are reduced, enabling the network to converge quickly to a good solution.

*2) Pruning Part*

The pruning procedure during online training of our CD network is an adaptive process to alleviate the computation complexity during training (and inference) by reducing network parameters while maintaining CD performance. Our OAIP utilizes the L1-norm-based filter-level pruning strategy due to its simplicity and effectiveness [20]. The flow chart of the pruning part in our OAIP is shown in Fig. 5. To mitigate performance decline resulting from pruning, OAIP adaptively determines how many filters need to be pruned and whether to perform the rolling back operation based on the check of increase of the loss values in Eq. (7). The rolling back operation is to revert both the network structure and parameters to the previous iteration. Note that the output of *Conv* 3-3 remains constant with the same input training samples during the training process due to the frozen weights of *Conv* 1-1 to *Conv* 3-3 in Fig. 2. Therefore, after the first forward propagation, we execute subsequent forward propagation directly from *Conv* 3-4 using the stored output of *Conv* 3-3.

OAIP prunes filters in *Conv* 3-4 to *Conv* 5-4 to achieve accelerated computation and reduced memory usage. In the pruning process, we set the pruning iteration interval as $P$, the maximum pruning rate in each pruning operation as $r^{\max}$, the number of filters of layer $l \in \Theta_p = \{Conv\ 3\text{-}4, Conv\ 4\text{-}1, \ldots, Conv\ 5\text{-}4\}$ to be pruned as $N^l$, and $N^l$ is initialized to the $r^{max} \times N_e^l$, where $N_e^l$ is the number of existing filters in layer $l$. In each pruning operation, the L1-norm sum of weights in each filter pair located symmetrically in two subnetworks $\{t_1, t_2\}$ is first calculated as its significance score,

$$S_c^l = \left\| \mathbf{F}_c^{l,t_1} \right\| + \left\| \mathbf{F}_c^{l,t_2} \right\|, \quad c \in \{1, \ldots, C_o^l\}, \qquad (28)$$

where $\mathbf{F}_c^{l,t_p}$ represents the $c$th filter of the weight matrix in layer $l$ of subnetwork $t_p$ ($p = 1, 2$) and $C_o^l$ is the number of output channels of layer $l$. Then, for each layer, $N^l$ filter pairs with the smallest values of $S_c^l$ are pruned.

OAIP prunes filter pairs with a period of pruning iteration interval $P$. We check the increase of the loss after $P/2$ training iterations from the current pruning operation, where three observed values are taken into account, i.e., the maximum loss $\mathcal{L}^{max}$ from the beginning of iteration until the iteration to be checked, the loss before the last pruning $\mathcal{L}^{last}$, and the loss value $\mathcal{L}_k$ at the $k$th iteration (the iteration to be checked). Pruning crucial filters often causes an increase in loss values. Therefore, we set the condition for executing the rolling back operations as

$$\mathcal{L}^{max} - \mathcal{L}_k < \theta(\mathcal{L}^{max} - \mathcal{L}^{last}), \qquad (29)$$

where $0 < \theta < 1$ represents a tolerated percentage factor. If $\mathcal{L}^{max} - \mathcal{L}_k < \theta(\mathcal{L}^{max} - \mathcal{L}^{last})$ in Eq. (29) is satisfied, the pruning operation may have removed important filters, resulting in performance degradation and an increase in the loss function. In this case, the parameters and structure of the CD network are rolled back just before the last pruning and then an attempt to prune $N^l/2$ filter pairs is made to reduce important filters to be pruned. The failure of the condition in Eq. (29) implies that removed filters result in negligible performance loss and we can continue training and greedily attempt further pruning $N^l = r^{max} \times N_e^l$ filter pairs. When the number of iterations $k$ exceeds the total number of iterations $T$, we halt the training for heterogeneous CD. The completion of online training signifies that the CD network is adapted to the current input data.

### D. Inference of Change Detection Map

After OAIP-based training, we use the trained CD network to process the bitemporal images $\{I^{pre}, I^{post}\}$. First, $\{I^{pre}, I^{post}\}$ are input to the trained CD network. Then, the CD network compares FRN outputs by

$$D_m(x, y) = \frac{1}{C_m} \sum_{z=1}^{C_m} (\text{FRN}_{x,y,z}^{t_1,m,pre} - \text{FRN}_{x,y,z}^{t_2,m,post})^2, \qquad (30)$$

where $\text{FRN}_{x,y,z}^{t_p,m,\tau}$ is the output of the FRN from $\Theta$ of subnetwork $t_p$ ($p = 1, 2$) with the input as $I^\tau$ ($\tau \in \{pre, post\}$). The difference map $I_d$ is then calculated based on the weighted sum of $\{D_m, m = 3, 4, 5\}$

$$I_d = \alpha_3 D_3 + \alpha_4 U(D_4) + \alpha_5 U(D_5), \qquad (31)$$

where $U$ is a bilinear interpolation function that upscales $D_4$ and $D_5$ to the same size with $D_3$. $\{\alpha_m, m = 3, 4, 5\}$ are consistent the same with those in Eqs. (8) and (9). Finally, the

Otsu [28] algorithm is employed to obtain a binary CD result $I_b$ based on $I_d$:

$$I_b = f_b(I_d) = \text{Otsu}(I_d). \qquad (32)$$

## V. EXPERIMENTAL RESULTS

In this section, experimental results are conducted to demonstrate the superiority of our proposed OAIP-based CD method for heterogeneous remote sensing images. First, we introduce three heterogeneous remote sensing image datasets and the training details. Then we conduct ablation experiments to analyze the effect of hyperparameter settings and demonstrate the contribution of each component of OAIP-based training to the lightweight CD. Finally, the performance of our proposed method is compared against four state-of-the-art heterogeneous CD methods, i.e., HPT [5], SCCN [8], X-Net [9], and S$^3$N combined with patch-level affinity matrix comparison training sample selection (Aff-S$^3$N) [12]. Experiments are conducted on a server equipped with Ubuntu 18 with an Intel® Xeon(R) Silver 4210 CPU @ 2.20GHz × 40, DDR4 RAM of 128GB, and an NVIDIA GeForce RTX 4090. The codes are implemented in Python 3.9 and PyTorch 1.13.

The accuracy $R_a$, the precision $R_p$, the recall rate $R_r$, and the kappa $Ka$, are used to evaluate the CD performance, where

$$R_a = \frac{n_a + n_c}{N_t}, R_p = \frac{n_a}{N_d}, R_r = \frac{n_a}{N_c}, Ka = \frac{R_a - p_e}{1 - p_e}. \qquad (33)$$

In Eq. (33), $n_a$ and $n_c$ are the number of changed and unchanged pixels that are correctly detected, respectively. $N_d$, $N_c$, and $N_t$ are the number of pixels detected as change, the number of truly changed pixels, and the total number of pixels, respectively. $p_e$ is the hypothetical probability of random agreements [28]. Note that for the case with sparse change areas, the kappa value is a more effective evaluation indicator than accuracy. In addition, running time and memory usage are used as efficiency measurements of the CD methods. The running time is the time used for the whole CD process, including training sample selection, online training of the CD network, and network inference. Memory, such as RAM that is utilized for rapid data access and temporary tasks is typically more constrained compared with storage in practice. Our research evaluates the average memory usage of the whole CD process to test the efficiency of CD methods.

### A. Datasets

*1) Gloucester Flood:* Fig. 6 shows remote sensing images, monitoring the flood that happened in Gloucester, U.K. on October 21, 1999. Fig. 6(a) is the pre-event image $I^{pre}$ derived by the SPOT satellite in 1999 and Fig. 6(b) is the post-event normalized difference vegetation index (NDVI) image $I^{post}$ in 2000. Both images have $2359 \times 1318$ pixels. The ground truth in Fig. 6(c) is from [5]. The post-image is duplicated into three channels to facilitate the training of the CD network.

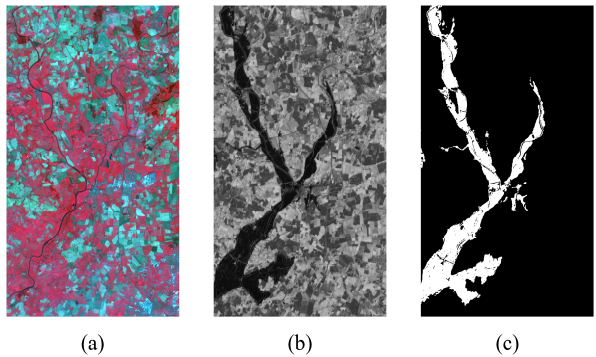

Fig. 6. Flood in Gloucester. (a) SPOT pre-event image. (b) NDVI post-event image. (c) Ground truth.

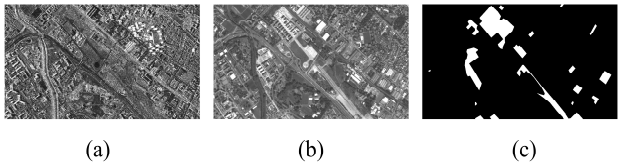

Fig. 7. Construction in Toulouse. (a) TerraSAR-X pre-event image. (b) Pleiades post-event image. (c) Ground truth.

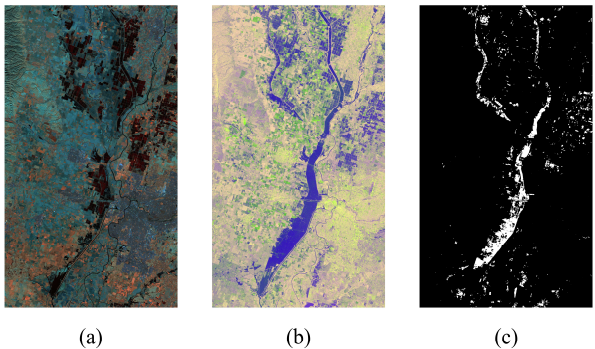

Fig. 8. Flood in California. (a) Landsat pre-event image. (b) Sentinel-1A post-event image. (c) Ground truth.

*2) Construction in Toulouse:* The dataset regarding construction in Toulouse is depicted in Fig. 7. Fig. 7(a) shows the SAR image from TerraSAR-X satellite in 2009 covering Toulouse, France. Fig. 7(b) is an optical image acquired by the Pleiades satellite in 2013. The size of Figs. 7(a)-(b) is $4400 \times 2600$ pixels. Both images are replicated into three channels before the CD processing. Fig. 7(c) depicts the ground truth of the change areas in Figs. 7(a) and (b) [31].

*3) Flood in California:* The dataset regarding flood in California is shown in Fig. 8. The optical image in Fig. 8(a) and the co-registered SAR image in Fig. 8(b) were acquired by Landsat 8 on 5 January 2017 and Sentinel-1A on 18 February 2017, respectively. The number of input channels of our CD network is 3, so we choose the 4th, 5th, and 6th channels in the optical image as the input to train the network. The SAR image is recorded in VV and VH polarizations. The result of dividing VV by VH is placed in the third channel for image augmentation. The size of both images in Figs. 8(a) and (b) is $3500 \times 2000$ pixels. The work in [6] provides the ground truth of the change areas in Figs. 8(a) and (b), as shown in Fig. 8(c).

### B. Training Details

The batch gradient descent (BGD) [27] is employed to train our CD network. For the selection of training samples, we set patch size to $64 \times 64$ for the dataset in Fig. 6. Due to the larger size of datasets in Figs. 7-8, which encompasses more features, the patch size is $128 \times 128$ to provide more information for network training. The tolerated percentage factor $\theta$ is set to 0.7. The $b_{gw}$ is fixed as 5 to ensure that scaling in Eq. (27) will not result in significant errors. The number of the total training iterations $T$ is set to 1000.

### C. Hyperparameter Setting

The number of training samples is significant for the online training of the CD network. The kappa values of our proposed method trained with different values of $n_h$ and $n_l$ on the dataset in Fig. 8 are shown in Fig. 9(a). The leftmost subplot of Fig. 9(a) shows the kappa values with the fixed $n_h = 20$ and different values of $n_l$. The rightmost subplot of Fig. 9(a) displays the kappa values with a fixed $n_l = 30$ and varying values of $n_h$. As shown in Fig. 9(a), the kappa values do not exhibit drastic fluctuations and demonstrate good robustness across different values of $n_l$ and $n_h$. As shown in Fig. 9(b), running times tested on different $\{n_l, n_h\}$ settings show small differences and memory usages slightly increase with the rise of the total batch size $n_l + n_h$. Without loss of generality, we use the setting of $\{n_l = 30, n_h = 20\}$ in subsequent experiments due to its good CD performance and efficiency.

Fig. 10(a) shows the effect of pruning iteration interval $P$ and the maximum pruning rate $r^{\max}$ on the CD performance and efficiency. The leftmost subplot of Fig. 10(a) illustrates kappa values of the CD results with a constant $P = 20$ and varying values of $r^{\max}$. The rightmost subplot showcases kappa values with a consistent $r^{\max} = 1/16$ and different values of $P$. As shown in Fig. 10(a), the kappa values show little variation across $r^{\max} \in [1/64, 1/4]$ and $P \in [6, 40]$, indicating good robustness of the CD performance to these hyperparameter settings. In Fig. 10(b), we show the running time (upper part) and memory usage (lower part) of the OAIP-based training process with different values of $P$ and $r^{\max}$. We can see that the higher values of $r^{\max}$ and the smaller values of $P$ lead to faster online training process and lower memory usage. Without loss of generality, the setting of $\{r^{\max} = 1/16, P = 20\}$ is selected for subsequent experiments.

### D. Contribution of All-Integer-Training and Pruning

Next, we investigate the contribution of all-integer training and pruning in the OAIP-based method to the lightweight CD. We compare the kappa values and resource consumption of online all-floating-point (OAF) training without all-integer

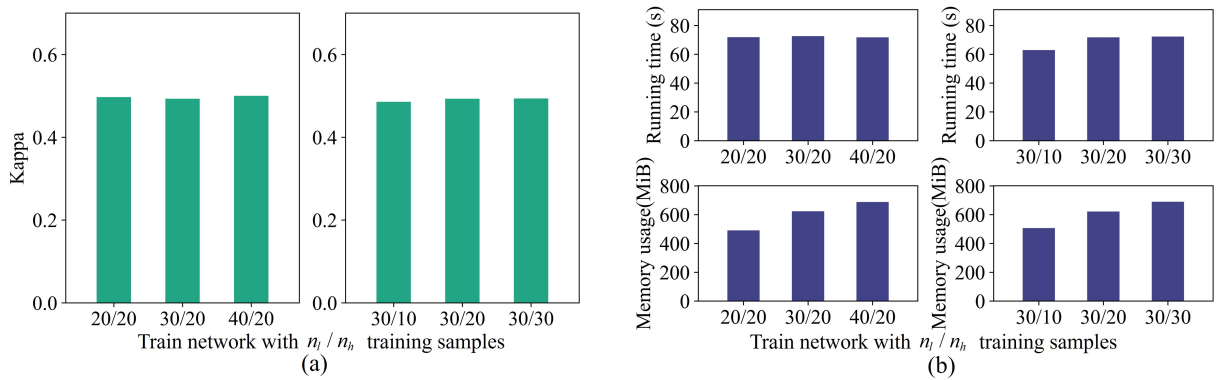

Fig. 9. Comparison of CD performance (a), running time [upper part of (b)], and memory usage [lower part of (b)] with different numbers of negative training samples $n_l$ and positive training samples $n_h$ on the dataset in Fig. 8.

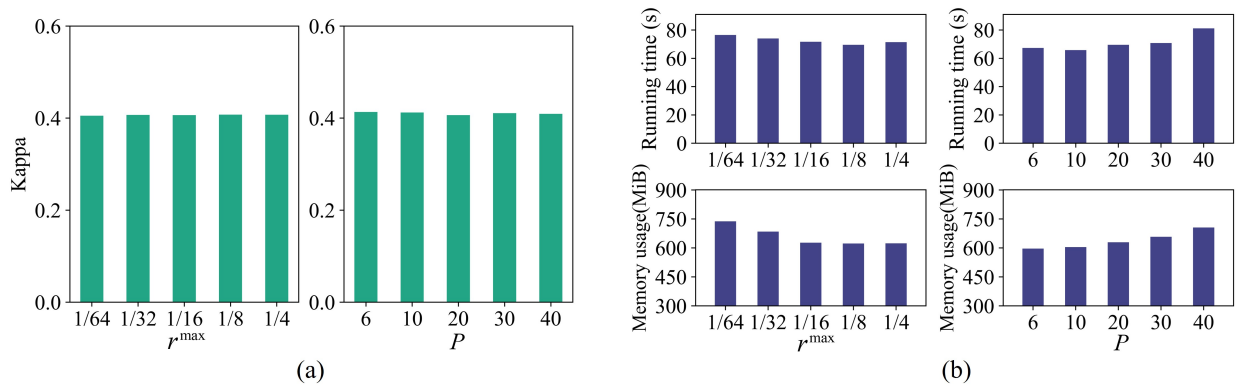

Fig. 10. Comparison of detection performance (a), running time [upper part of (b)], and memory usage [lower part of (b)] with different combinations of $r^{\max}$ and $P$ on the dataset in Fig. 7.

## TABLE I
## Ablation Experiments Illustrating the Contribution of Each Component to the Lightweight Effect of Our Method (without Restricted Memory Usage)

| Training Strategies | Dataset in Fig. 6 | | | Dataset in Fig. 7 | | | Dataset in Fig. 8 | | |
|---|---|---|---|---|---|---|---|---|---|
| | $Ka$ | Running time | Memory usage | $Ka$ | Running Time | Memory usage | $Ka$ | Running Time | Memory usage |
| OAF | 0.8110 | 74s | 825.2 MiB | 0.3410 | 173s | 2162.8 MiB | 0.4668 | 174s | 2045.5 MiB |
| OAI | 0.7441 | **51s** | 372.1 MiB | **0.4164** | 79s | 908.5 MiB | 0.4620 | 79s | 874.1 MiB |
| OAIP | **0.8114** | 52s | **228.3 MiB** | 0.4066 | **68s** | **629.5 MiB** | **0.4932** | **71s** | **622.2 MiB** |

The **boldface** indicates the best results. The OAF, and OAI represent the online all-float training, and online all-integer training, respectively.

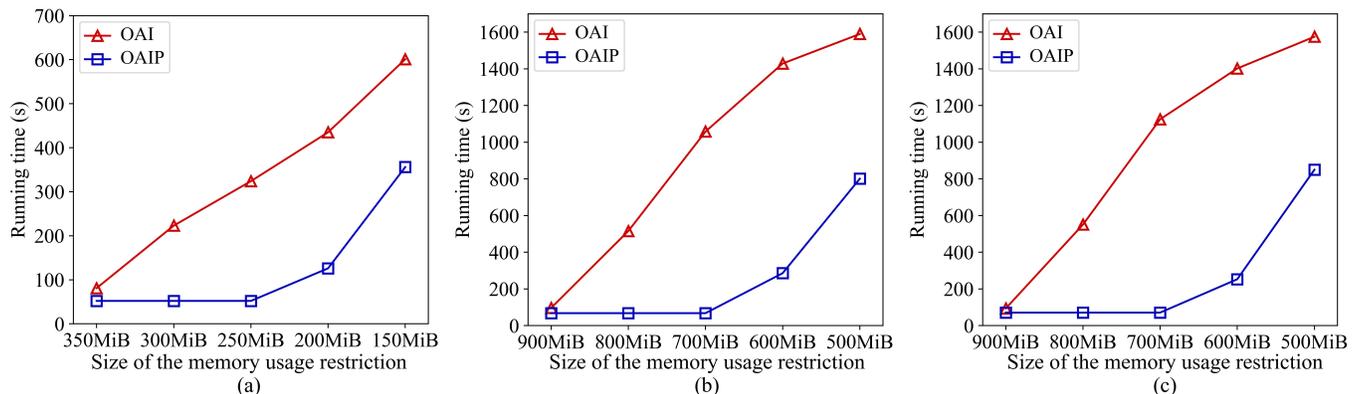

Fig. 11. Running times of the whole CD process under the condition of restricted memory usage. (a), (b), and (c) are the results tested on datasets in Figs. 6, 7, and 8, respectively.

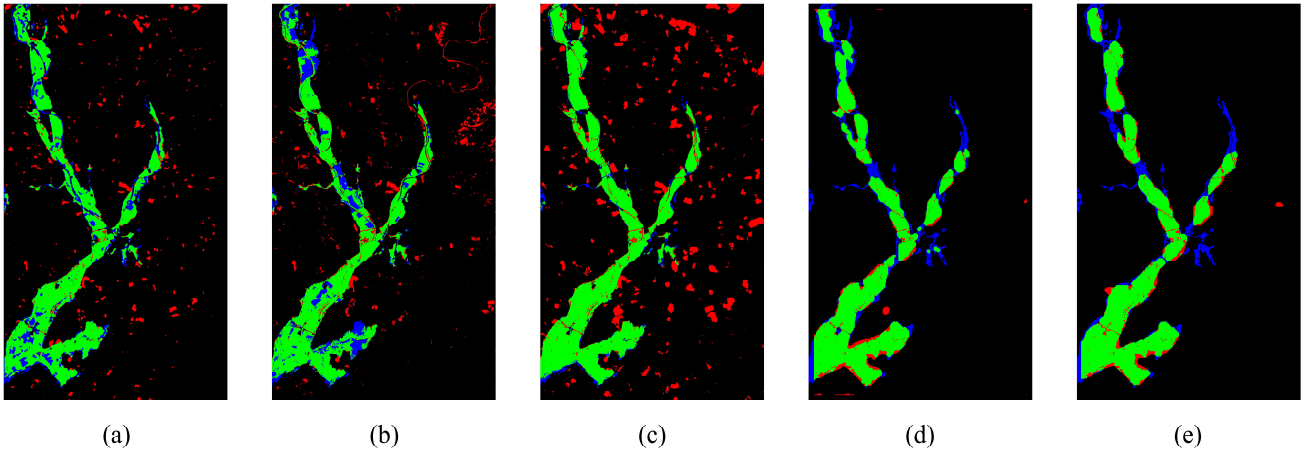

Fig. 12. Comparison results using the dataset in Fig. 6, achieved by (a) HPT, (b) SCCN, (c) X-Net, (d) Aff-S³N, and (e) our proposed method. The green, red, blue, and black colors are the areas of true positives, false alarms, missed targets, and true negatives, respectively.

### TABLE II
### COMPARISON OF CD METHODS ON THE DATASET IN FIG. 6

| Methods | $R_a$ | $R_p$ | $R_r$ | $Ka$ | Running time | Memory usage |
|---------|-------|-------|-------|------|--------------|--------------|
| HPT [5] | 0.9522 | 0.8030 | 0.7929 | 0.7708 | 899s | 1780.3 MiB |
| SCCN [8] | 0.9456 | 0.7729 | 0.7696 | 0.7404 | 17573s | 724.2 MiB |
| X-Net [9] | 0.9380 | 0.6747 | **0.9254** | 0.7454 | 476s | 557.0 MiB |
| Aff-S³N [12] | **0.9612** | **0.8958** | 0.7781 | 0.8110 | 74s | 825.2 MiB |
| Proposed Method | 0.9610 | 0.8854 | 0.7873 | **0.8114** | **52s** | **228.3 MiB** |

The **boldface** indicates the best results.

calculation framework and pruning, online all-integer (OAI) training without pruning, and our proposed OAIP in Table I. In Table I, OAI shows a significant reduction in terms of running times and memory usages in comparison with OAF, which exhibits the crucial efficiency of all-integer training. Compared with OAI, our proposed OAIP does not show obvious degradation for kappa values of the CD results, but the memory usages and running times are notably decreased for datasets in Figs. 7 and 8, demonstrating the efficiency and effectiveness of the pruning. Note that OAIP shows a slight increase in running time compared with that of OAI when processing the dataset in Fig. 6, but memory usage decreases significantly. To more fairly indicate the effect of the pruning through the comparison of the lightweight effects of OAI and OAIP, we conducted experiments under conditions of restricted memory usage. Restricted memory scenarios usually occur in edge-computing devices, such as those found in nanosatellites. Restricting memory usage is meaningful even when memory is abundant, as it allows for the saved memory to be allocated for training other networks. Fig. 11 depicts the running times of OAI and OAIP under memory-restricted conditions on datasets in Figs. 6-8. When both methods are constrained within the same memory-restricted scenario, OAIP exhibits much shorter running times than OAI, presenting the efficiency of the online adaptive pruning.

### E. Comparison with Existing Competitors

Fig. 12 depicts the CD results of HPT [5], SCCN [8], X-Net [9], Aff-S³N [12], and our proposed OAIP-based CD method when testing the heterogeneous image dataset in Fig. 6. Compared with our proposed OAIP-based method, HPT [5], SCCN [8], X-Net [9] show more false alarms in CD results. This result is attributed to their limitations in effectively understanding semantic information. The result of the CD performance of Aff-S³N [12] is similar to that of the proposed method as their basic network architectures are closely related. Table II shows the quantitative evaluations using the dataset in Fig. 6. Although the HPT [5] and X-Net [9] achieve higher $R_r$ values, their $R_p$ values are comparatively lower. This is because HPT [5] and X-Net [9] tend to learn low-level features that may be degraded by image noise. Aff-S³N [12] and our OAIP-based method achieve higher comprehensive measurements $R_a$ and $Ka$, since they mainly exploit the deep-level features of the heterogeneous remote sensing images. The CD performance of our proposed method is similar to that of Aff-S³N [12], indicating that the all-integer calculation framework and adaptive pruning strategy do not lead to noticeable degradation in terms of CD performance. Table II also shows that our proposed method outperforms HPT [5], SCCN [8], X-Net [9], and Aff-S³N [12] in terms of running time and memory usage during training. For example,

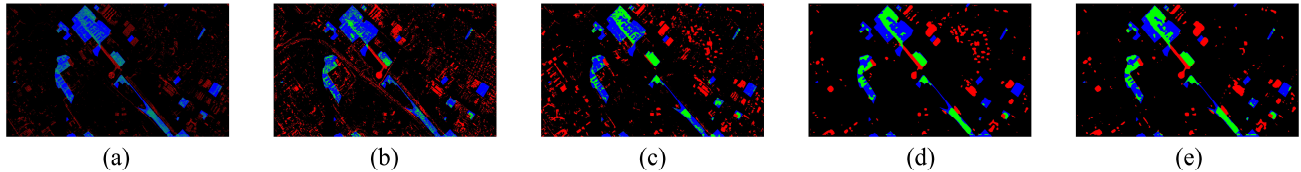

Fig. 13. Comparison results using the dataset in Fig. 7, achieved by (a) HPT, (b) SCCN, (c) X-Net, (d) Aff-S³N, and (e) our proposed method. The green, red, blue, and black colors are the areas of true positives, false alarms, missed targets, and true negatives, respectively.

TABLE III
COMPARISON OF CD METHODS ON THE DATASET IN FIG. 7

| Methods | $R_a$ | $R_p$ | $R_r$ | $Ka$ | Running time | Memory usage |
|---|---|---|---|---|---|---|
| HPT [5] | 0.9061 | 0.3594 | 0.2196 | 0.2256 | 2179s | 8023.0 MiB |
| SCCN [8] | 0.8622 | 0.2011 | 0.2424 | 0.1405 | 41547s | 2666.8 MiB |
| X-Net [9] | 0.8806 | 0.2894 | 0.3372 | 0.2465 | 903s | 2226.6 MiB |
| Aff-S³N [12] | 0.9081 | 0.4168 | 0.3671 | 0.3410 | 173s | 2162.8 MiB |
| Proposed Method | **0.9155** | **0.4703** | **0.4356** | **0.4066** | **68s** | **629.5 MiB** |

The **boldface** indicates the best results.

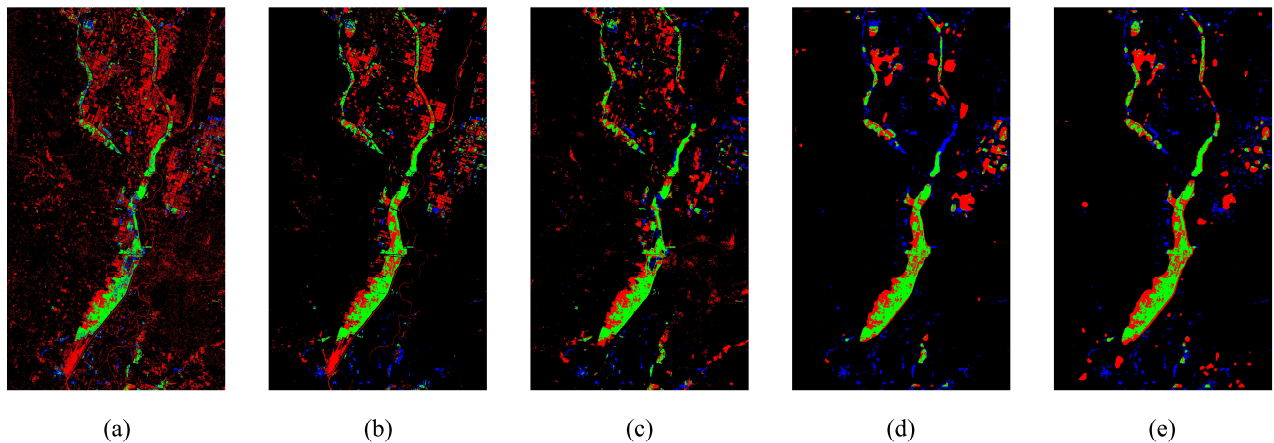

Fig. 14. Comparison results using the dataset in Fig. 8, achieved by (a) HPT, (b) SCCN, (c) X-Net, (d) Aff-S³N, and (e) our proposed method. The green, red, blue, and black colors are the areas of true positives, false alarms, missed targets, and true negatives, respectively.

TABLE IV
COMPARISON OF CD METHODS ON THE DATASET IN FIG. 8

| Methods | $R_a$ | $R_p$ | $R_r$ | $Ka$ | Running time | Memory usage |
|---|---|---|---|---|---|---|
| HPT [5] | 0.8534 | 0.1838 | 0.6845 | 0.2373 | 1283s | 4540.5 MiB |
| SCCN [8] | 0.9294 | 0.3568 | **0.7648** | 0.4541 | 23443s | 1474.2 MiB |
| X-Net [9] | 0.9335 | 0.3683 | 0.7269 | 0.4574 | 638s | 2188.8 MiB |
| Aff-S³N [12] | 0.9409 | 0.4290 | 0.5915 | 0.4668 | 174s | 2045.5 MiB |
| Proposed Method | **0.9426** | **0.4439** | 0.6354 | **0.4932** | **71s** | **622.2 MiB** |

The **boldface** indicates the best results.

compared with Aff-S$^3$N, our proposed method reduces training time by 30% and memory usage by 72%, thanks to the excellent computational efficiency of all-integer training and adaptive pruning.

The binary CD results and quantitative evaluations of HPT [5], SCCN [8], X-Net [9], Aff-S$^3$N [12], and the proposed method regarding the datasets in Fig. 7 and Fig. 8 are depicted in Figs. 13, 14 and Tables III, IV, respectively. Due to the larger size of the datasets in Figs. 7 and 8 compared to the dataset in Fig. 6, both results require more running time and memory usage when processing Figs. 7 and 8 than Fig. 6. Similar to Fig. 12 and Table II, we can observe from Figs. 13 and 14 and Tables III and IV that our method outperforms HPT [5], SCCN [8], X-Net [9], and Aff-S$^3$N [12] in terms of running time and memory usage, and does not show notable deterioration of CD performance. Again, the superiority of our proposed OAIP-based method is demonstrated.

## VI. CONCLUSION

In this work, we introduce a new OAIP-based CD method for heterogeneous remote sensing images that improves the efficiency of existing online training-based methods. We adopt a patch-based affinity matrix comparison strategy to adaptively (and online) select training samples for our CD network, i.e., the Siamese network with the VGG structure. Then, two new operations are designed to lighten the online training process of our CD network. First, an all-integer training framework is employed to convert floating-point calculations to all-integer calculations to accelerate the computation and reduce memory usage. PLBESS is exploited to reduce the calculation errors caused by all-integer training. Second, adaptive filter-level pruning is utilized to remove unimportant filters to lighten the online training process. Ablation experiments demonstrate the contribution of all-integer training and pruning to the lightweight effect of our method. We also show the robustness of the CD performance of our proposed method in terms of hyperparameter settings. Finally, experimental results show that the newly proposed OAIP-based CD method achieves similar or slightly better CD performance with fewer running time and memory usage in comparison with existing state-of-the-art heterogeneous CD methods.

In the future, a feasible direction is to explore the use of lower bit precision calculations, such as 4-bit integer or 2-bit integer, to achieve improved CD efficiency with maintained CD accuracy.

## REFERENCES

[1] Q. Xu, Y. Shi, J. Guo, C. Ouyang, and X. X. Zhu, "UCDFormer: Unsupervised change detection using a transformer-driven image translation," *IEEE Trans. Geosci. Remote Sens.*, vol. 61, Sep. 2023, Art. no. 5619917.

[2] M. Zhang, Z. Liu, J. Feng, L. Liu, and L. Jiao, "Remote sensing image change detection based on deep multi-scale multi-attention Siamese transformer network," *Remote Sens.*, vol. 15, no. 3, 2023, Art. no. 842.

[3] L. Kondmann, S. Saha and X. X. Zhu, "SemiSiROC: Semisupervised Change Detection With Optical Imagery and an Unsupervised Teacher Model," *IEEE J. Sel. Topics Appl. Earth Observ. Remote Sens.*, vol. 16, pp. 3879-3891, Apr. 2023.

[4] M. G. Gong, P. Z. Zhang, L. Su, and J. Liu, "Coupled dictionary learning for change detection from multisource data," *IEEE Trans. Geosci. Remote Sens.*, vol. 54, no. 12, pp. 7077–7091, Dec. 2016.

[5] Z. Liu, G. Li, G. Mercier, Y. He, and Q. Pan, "Change detection in heterogeneous remote sensing images via homogeneous pixel transformation," *IEEE Trans. Image Process.*, vol. 27, no. 4, pp. 1822-1834, Apr. 2018.

[6] L. T. Luppino, F. M. Bianchi, G. Moser, and S. N. Anfinsen, "Unsupervised image regression for heterogeneous change detection," *IEEE Trans. Geosci. Remote Sens.*, vol. 57, no. 12, pp. 9960–9975, Dec. 2019.

[7] Y. Sun, L. Lei, X. Li, H. Sun, and G. Kuang, "Nonlocal patch similarity based heterogeneous remote sensing change detection," *Pattern Recognit.*, vol. 109, Jan. 2021, Art. no. 107598.

[8] J. Liu, M. Gong, K. Qin, and P. Zhang, "A deep convolutional coupling network for change detection based on heterogeneous optical and radar images," *IEEE Trans. Neural Netw. Learn. Syst.*, vol. 29, no. 3, pp. 545–559, Mar. 2016.

[9] L. T. Luppino *et al.*, "Deep image translation with an affinity-based change prior for unsupervised multimodal change detection," *IEEE Trans. Geosci. Remote Sens.*, vol. 60, pp. 1–22, Feb. 2022.

[10] L. T. Luppino *et al.*, "Code-aligned autoencoders for unsupervised change detection in multimodal remote sensing images," *IEEE Trans. Neural Netw. Learn.*, vol. 35, no. 1, pp. 60–72, Jan. 2024.

[11] X. Jiang, G. Li, Y. Liu, X.-P. Zhang, and Y. He, "Change detection in heterogeneous optical and SAR remote sensing images via deep homogeneous feature fusion," *IEEE J. Sel. Topics Appl. Earth Observ. Remote Sens.*, vol. 13, pp. 1551–1566, Apr. 2020.

[12] X. Jiang, G. Li, X.-P. Zhang, and Y. He, "A semisupervised Siamese network for efficient change detection in heterogeneous remote sensing images," *IEEE Trans. Geosci. Remote Sens.*, vol. 60, pp. 1–18, Mar. 2021.

[13] C. Zhu, S. Han, H. Mao, and W. J. Dally, "Trained ternary quantization," 2016, *arXiv:1612.01064*.

[14] M. Courbariaux, I. Hubara, D. Soudry *et al.*, "Binarized neural networks: Training deep neural networks with weights and activations constrained to+ 1 or-1," 2016, *arXiv:1602.02830*.

[15] F. Zhu *et al.*, "Towards unified INT8 training for convolutional neural network," in *Proc. IEEE/CVF Conf. Comput. Vis. Pattern Recognit.*, Jun. 2020, pp. 1969–1979.

[16] D. Das *et al.*, "Mixed precision training of convolutional neural networks using integer operations," 2018, *arXiv: 1802.00930*.

[17] Y. Yang, L. Deng, S. Wu, T. Yan, Y. Xie, and G. Li, "Training high-performance and large-scale deep neural networks with full 8-bit integers," *Neural Netw.*, vol. 125, pp. 70–82, May. 2020.

[18] B. Liu, Y. Cai, Y. Guo, and X. Chen, "TransTailor: Pruning the pre-trained model for improved transfer learning," 2021, *arXiv:2103.01542*.

[19] Z. Liu, J. Li, Z. Shen, G. Huang, S. Yan, and C. Zhang, "Learning efficient convolutional networks through network slimming," in *Proc. Int. Conf. Comput. Vis.*, 2017, pp. 2755–2763.

[20] H. Li, A. Kadav, I. Durdanovic, H. Samet, and H. Graf, "Pruning filters for efficient convnets," in *Proc. Int. Conf. Learn. Represent.*, 2017, pp. 1–13.

[21] P. Singh, V. K. Verma, P. Rai, and V. P. Namboodiri, "Acceleration of deep convolutional neural networks using adaptive filter pruning," *IEEE Trans Nucl Sci*, vol. 14, no. 4, pp. 838–847, May. 2020.

[22] A. Kusupati, V. Ramanujan, R. Somani *et al.*, "Soft threshold weight reparameterization for learnable sparsity," in *Proc. Int. Conf. Mach. Learn.*, 2020, pp. 5544–5555

[23] M. Wang, S. Rasoulinezhad, P. H. W. Leong, and H. K.-H. So, "NITI: Training integer neural networks using integer-only arithmetic," *IEEE Trans. Parallel Distrib. Syst.*, vol. 33, no. 11, pp. 3249–3261, Nov. 2022.

[24] J. Zhang, X. Chen, M. Song, and T. Li, "Eager pruning: Algorithm and architecture support for fast training of deep neural networks," in *Proc. ACM/IEEE 46th Annu. Int. Symp. Comput. Architect.*, 2019, pp. 292–303.

[25] S. Singh and S. Krishnan, "Filter response normalization layer: Eliminating batch dependence in the training of deep neural networks," in *Proc. IEEE Conf. Comput. Vis. Pattern Recognit.*, 2020, pp. 11234–11243.

[26] J. Lu, C. Ni, and Z. Wang, "ETA: An efficient training accelerator for DNNs based on hardware-algorithm co-optimization," *IEEE Trans. Neural Netw. Learn. Syst.*, vol. 34, no. 10, pp. 7660–7674, Oct. 2023.

[27] S. Sun, Z. Cao, H. Zhu, and J. Zhao, "A survey of optimization methods from a machine learning perspective," *IEEE Trans. Cybern.*, vol. 50, no. 8, pp. 3668–3681, Aug. 2020.

[28] R. C. Gonzalez and R. E. Woods, *Digital Image Processing*, 3rd ed. New York, NY, USA: Pearson, 2007.

[29] A. Krizhevsky, "Learning multiple layers of features from tiny images," Univ. Toronto, Toronto, ON, Canada, Tech. Rep., 2009. [Online]. Available: https://www.cs.toronto.edu/~kriz/learning-features-2009-TR.pdf

[30] Z. Lv, H. Huang, L. Gao, J. A. Benediktsson, M. Zhao, and C. Shi, "Simple multiscale UNet for change detection with heterogeneous remote sensing images," *IEEE Geosci. Remote Sens. Lett.*, vol. 19, 2022, Art. no. 2504905.

[31] R. Touati, M. Mignotte, and M. Dahmane, "Multimodal change detection in remote sensing images using an unsupervised pixel pairwise-based Markov random field model," *IEEE Trans. Image Process.*, vol. 29, pp. 757–767, 2020.

[32] H. Mao *et al.*, "Exploring the granularity of sparsity in convolutional neural networks," in *Proc. IEEE Conf. Comput. Vis. Pattern Recognit. Workshops*, Jul. 2017, pp. 1927–1934.

[33] G. Yang et al., "Algorithm/hardware codesign for real-time on-satellite CNN-based ship detection in SAR imagery," *IEEE Trans. Geosci. Remote Sens.*, vol. 60, 2022, Art. no. 5226018.

[34] G. Gao, L. Yao, W. Li, L. Zhang, and M. Zhang, "Onboard information fusion for multisatellite collaborative observation: Summary, challenges, and perspectives," *IEEE Geosci. Remote Sens. Mag.*, vol. 11, no. 2, pp. 40–59, Jun. 2023.

[35] B. Zhang *et al.*, "Progress and challenges in intelligent remote sensing satellite systems," *IEEE J. Sel. Top. Appl. Earth Obs. Remote Sens.*, vol. 15, pp. 1814-1822, 2022.

[36] W. Zhao, Z. Wang, M. Gong, and J. Liu, "Discriminative feature learning for unsupervised change detection in heterogeneous images based on a coupled neural network," *IEEE Trans. Geosci. Remote Sens.*, vol. 55, no. 12, pp. 7066–7080, Dec. 2017.

[37] Z. Lv, J. Liu, W. Sun, T. Lei, J. A. Benediktsson, and X. Jia, "Hierarchical attention feature fusion-based network for land cover change detection with homogeneous and heterogeneous remote sensing images," *IEEE Trans. Geosci. Remote Sens.*, vol. 61, Nov. 2023.

[38] J. Wu *et al.*, "A multiscale graph convolutional network for change detection in homogeneous and heterogeneous remote sensing images," *Int. J. Appl. Earth Observ. Geoinf.*, vol. 105, Dec. 2021, Art. no. 102615.

[39] K. Jiang, J. Liu, F. Liu, W. Zhang, Y. Liu, and J. Shi, "Dual UNet: A novel Siamese network for change detection with cascade differential fusion," in *Proc. IEEE Int. Geosci. Remote Sens. Symp.*, Jul. 2022, pp. 1428–1431.

[40] R. C. Daudt, B. Le Saux, and A. Boulch, "Fully convolutional siamese networks for change detection," in *Proc. 25th IEEE Int. Conf. Image Process.*, Oct. 2018, pp. 4063–4067.

[41] J. Shi, T. Wu, A. K. Qin *et al.*, "Self-Guided Autoencoders for Unsupervised Change Detection in Heterogeneous Remote Sensing Images," *IEEE Trans. Artif. Intell.*, pp. 1-13, Jan. 2024.

[42] T. Han, Y. Tang, Y. Chen, "Heterogeneous Image Change Detection Based on Two-Stage Joint Feature Learning" in *Proc. IEEE Int. Geosci. Remote Sens. Symp.*, Sep. 2022, pp. 3215-3218.

[43] https://www.nvidia.com/en-us/geforce/ turing

[44] K. Simonyan, A. Zisserman, "Very deep convolutional networks for large-scale image recognition," 2014, *arXiv:1409.1556*.

[45] X. Lin, L. Yu, K. T. Cheng *et al.*, "The lighter the better: rethinking transformers in medical image segmentation through adaptive pruning," 2022, *arXiv:2206.14413*.

[46] H. Zhang *et al.*, "Optical and SAR image dense registration using a robust deep optical flow framework," *IEEE J. Sel. Topics Appl. Earth Observ. Remote Sens.*, vol. 16, pp. 1269–1294, 2023.

[47] L. Li, L. Han, M. Ding, and H. Cao, "Multimodal image fusion framework for end-to-end remote sensing image registration," *IEEE Trans. Geosci. Remote Sens.*, vol. 61, 2023, Art. no. 5607214.

[48] W. Li, L. Xue, X. Wang, and G. Li, "ConvTransNet: A CNN–transformer network for change detection with multiscale global–local representations," *IEEE Trans. Geosci. Remote Sens.*, vol. 61, 2023.

[49] W. Li, L. Xue, X. Wang, and G. Li, "MCTNet: A multi-scale CNN-transformer network for change detection in optical remote sensing images," in *Proc. IEEE 26th Int. Conf. Inf. Fusion*, 2023, pp. 1–5.

[50] C. Zhang, X. Wang, G. Li, "Lightweight change detection of heterogeneous remote sensing images based on online all-integer-pruning training," in *IGARSS 2024-2024 IEEE Int. Geosci. Remote Sens. Symp.* IEEE, 2024, (in press).